\documentclass{article}
\pdfoutput=1  % for arXiv to use pdflatex

\PassOptionsToPackage{numbers}{natbib}
\usepackage[final]{nips2016}

\usepackage[utf8]{inputenc} % allow utf-8 input
\usepackage[T1]{fontenc}    % use 8-bit T1 fonts
\usepackage[english]{babel}
\usepackage[colorlinks=true,allcolors=blue]{hyperref}       % hyperlinks
\usepackage{amsfonts}       % blackboard math symbols
\usepackage{amssymb}
\usepackage{amsmath}
\usepackage{xfrac}       % compact symbols for 1/2, etc.
\usepackage{subfigure}
\usepackage{graphicx}
\graphicspath{{figures/}}
\usepackage{booktabs}
\usepackage{setspace}
\usepackage{color}
\usepackage{capt-of}

\usepackage{titlesec}		% space before / after headings
\titlespacing\section{0pt}{3.3pt plus 1pt minus 2pt}{1.8pt plus 1pt minus 1pt}
\titlespacing\subsection{0pt}{3.3pt plus 1pt minus 2pt}{2.7pt plus 1pt minus 2pt}
\usepackage{parskip} 

\title{Convolutional Neural Networks on Graphs\\
with Fast Localized Spectral Filtering\\}

\author{
	Michaël Defferrard \hspace{1cm} Xavier Bresson \hspace{1cm} Pierre Vandergheynst %\thanks{further info} \\
	\vspace{0.2cm} \\
	EPFL, Lausanne, Switzerland \\
	\texttt{\{michael.defferrard,xavier.bresson,pierre.vandergheynst\}@epfl.ch} \\
}

\DeclareMathOperator*{\diag}{diag}

\DeclareMathOperator*{\spn}{span}
\newcommand{\G}{\mathcal{G}}
\newcommand{\V}{\mathcal{V}}
\newcommand{\E}{\mathcal{E}}
\newcommand{\bO}{\mathcal{O}}
\newcommand{\R}{\mathbb{R}}
\newcommand{\figref}[1]{Figure~\ref{fig:#1}}
\newcommand{\tabref}[1]{Table~\ref{tab:#1}}
\newcommand{\secref}[1]{Section~\ref{sec:#1}}
\newcommand{\eqnref}[1]{(\ref{eq:#1})}

\begin{document}

\maketitle

\begin{abstract}

In this work, we are interested in generalizing convolutional neural networks
(CNNs) from low-dimensional regular grids, where image, video and speech are
represented, to high-dimensional irregular domains, such as social networks,
brain connectomes or words' embedding, represented by graphs.
We present a formulation of CNNs in the context of spectral graph theory, which
provides the necessary mathematical background and efficient numerical schemes
to design fast localized convolutional filters on graphs. Importantly, the
proposed technique offers the same linear computational complexity and constant
learning complexity as classical CNNs, while being universal to any graph
structure. Experiments on MNIST and 20NEWS demonstrate the ability of this
novel deep learning system to learn local, stationary, and compositional
features on graphs.

	%there is nothing to lose by going to the graph domain
	%except the rotation invariance
	
	%not a generalization because it provides more invariance, but we don't want
	%to say it.
	%Our formulation is not a direct generalization of classical CNNs, in the
	%sense that it is not a spatial construction.
\end{abstract}

%Previous works have been developed along this today essential line of
%research, but they lack of mathematical foundations and efficient learning of
%filters.

\section{Introduction}

%Convolutional neural networks (CNNs) have greatly improved state-of-the-art performances in a number of fields, notably computer vision. 

	%Graphs can either be the natural domain of the data, such as social networks,
	%biological graphs like gene regulatory and brain connectivity networks,
	%telecommunication networks, or constructed from the data, such as similarity
	%graphs.
% Constructed graphs make sense because of comp complexity, as evidenced by LRF.

Convolutional neural networks \cite{pro:LeCunBottouBengioHaffner98MNIST} offer
an efficient architecture to extract highly meaningful statistical patterns in
large-scale and high-dimensional datasets. The ability of CNNs to learn local
stationary structures and compose them to form multi-scale hierarchical
patterns has led to breakthroughs in image, video, and sound recognition tasks
\cite{art:LeCunBengioHinton15DL}. Precisely, CNNs extract the local
stationarity property of the input data or signals by revealing local features
that are shared across the data domain. These similar features are identified
with localized convolutional filters or kernels, which are learned from the
data. Convolutional filters are shift- or translation-invariant filters,
meaning they are able to recognize identical features independently of their
spatial locations. Localized kernels or compactly supported filters refer to
filters that extract local features independently of the input data size, with
a support size that can be much smaller than the input size.

User data on social networks, gene data on biological regulatory networks, log
data on telecommunication networks, or text documents on word embeddings are
important examples of data lying on irregular or non-Euclidean domains that
can be structured with graphs, which are universal representations of
heterogeneous pairwise relationships. Graphs can encode complex geometric
structures and can be studied with strong mathematical tools such as spectral
graph theory \cite{book:Chung97Spectral}.

A generalization of CNNs to graphs is not straightforward as the convolution
and pooling operators are only defined for regular grids. This makes this
extension challenging, both theoretically and implementation-wise. The major
bottleneck of generalizing CNNs to graphs, and one of the primary goals of this
work, is the definition of localized graph filters which are efficient to
evaluate and learn.
Precisely, the main contributions of this work are summarized below.
\begin{enumerate}
\item \textbf{Spectral formulation.} A spectral graph theoretical formulation
	of CNNs on graphs built on established tools in graph signal processing (GSP).
	\cite{art:ShumanNarangFrossardOrtegaVandergheynst13ReviewSPG}.
\item \textbf{Strictly localized filters.} Enhancing
	\cite{art:BrunaZarembaSzlamLeCun13DLgraphs}, the proposed spectral filters
	are provable to be strictly localized in a ball of radius $K$, i.e. $K$
	hops from the central vertex.  
\item \textbf{Low computational complexity.} The evaluation complexity of our
	filters is linear w.r.t. the filters support's size $K$ and the number of
	edges $|\E|$. Importantly, as most real-world graphs are highly sparse, we
	have $|\E| \ll n^2$ and $|\E| = kn$ for the widespread $k$-nearest neighbor
	(NN) graphs, leading to a linear complexity w.r.t the input data size $n$.
	Moreover, this method avoids the Fourier basis altogether, thus the
	expensive eigenvalue decomposition (EVD) necessary to compute it as well as
	the need to store the basis, a matrix of size $n^2$. That is especially
	relevant when working with limited GPU memory. Besides the data, our method
	only requires to store the Laplacian, a sparse matrix of $|\E|$ non-zero
	values.
\item \textbf{Efficient pooling.} We propose an efficient pooling strategy on
	graphs which, after a rearrangement of the vertices as a binary tree
	structure, is analog to pooling of 1D signals.
\item \textbf{Experimental results.} We present multiple experiments that
	ultimately show that our formulation is (i) a useful model, (ii)
	computationally efficient and (iii) superior both in accuracy and
	complexity to the pioneer spectral graph CNN introduced in
	\cite{art:BrunaZarembaSzlamLeCun13DLgraphs}. We also show that our graph
	formulation performs similarly to a classical CNNs on MNIST and study the
	impact of various graph constructions on performance. The
	TensorFlow \cite{abadi_tensorflow_2016} code to reproduce our results and
	apply the model to other data is available as an open-source
	software.\footnote{\url{https://github.com/mdeff/cnn_graph}}
\end{enumerate}

\begin{figure}[t]
	\centering
	\includegraphics[width=\textwidth]{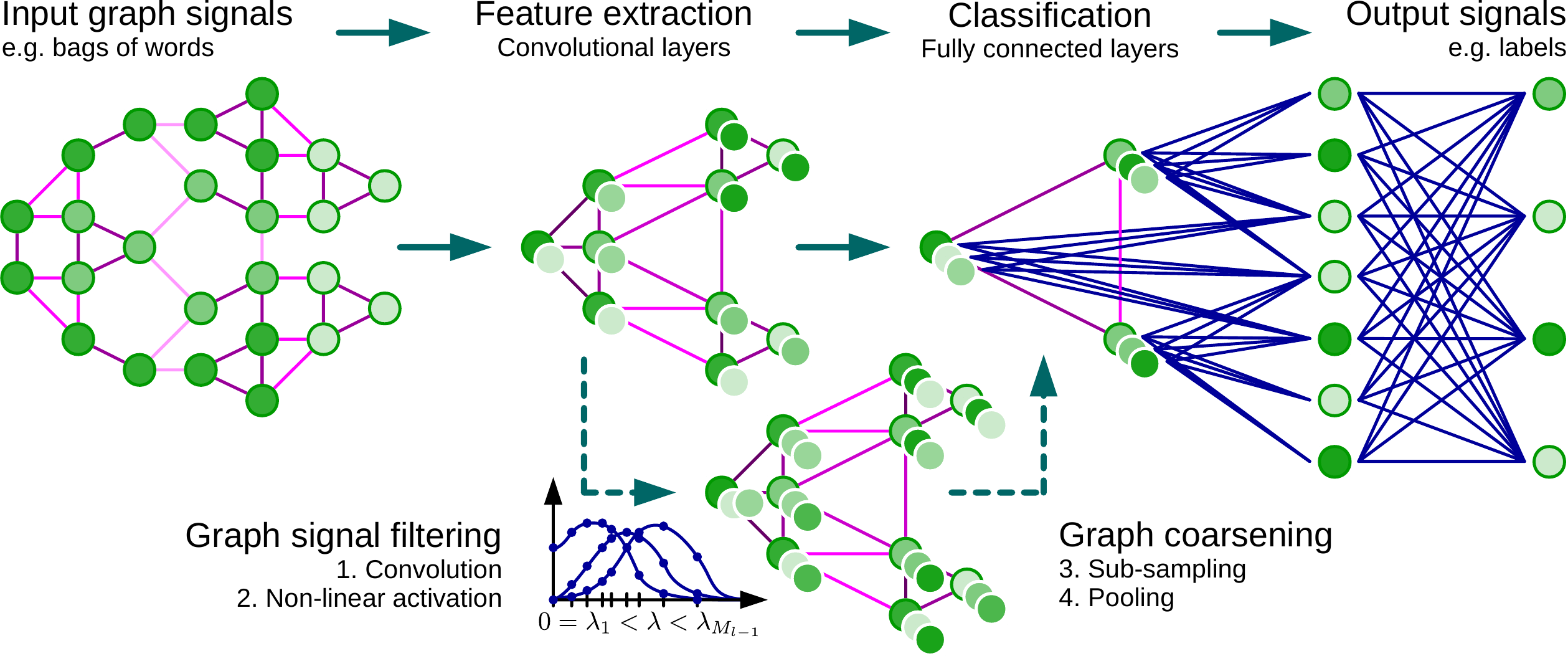}
	\caption{Architecture of a CNN on graphs and the four ingredients of a (graph) convolutional layer.}
	\label{fig:architecture}
\end{figure}

\section{Proposed Technique}

Generalizing CNNs to graphs requires three fundamental steps: (i) the design of
localized convolutional filters on graphs, (ii) a graph coarsening procedure
that groups together similar vertices and (iii) a graph pooling operation that
trades spatial resolution for higher filter resolution. 

\subsection{Learning Fast Localized Spectral Filters} \label{sec:filters}

There are two strategies to define convolutional filters; either from a spatial
approach or from a spectral approach. By construction, spatial approaches
provide filter localization via the finite size of the kernel. However,
although graph convolution in the spatial domain is conceivable, it
faces the challenge of matching local neighborhoods, as pointed out in
\cite{art:BrunaZarembaSzlamLeCun13DLgraphs}. Consequently, there is no unique
mathematical definition of translation on graphs from a spatial perspective. On
the other side, a spectral approach provides a well-defined localization
operator on graphs via convolutions with a Kronecker delta implemented in the
spectral domain \cite{art:ShumanNarangFrossardOrtegaVandergheynst13ReviewSPG}.
The convolution theorem \cite{book:Mallat99wavelets} defines convolutions as
linear operators that diagonalize in the Fourier basis (represented by the
eigenvectors of the Laplacian operator). However, a filter defined in the
spectral domain is not naturally localized and translations are costly due to
the $\bO(n^2)$ multiplication with the graph Fourier basis. Both limitations
can however be overcome with a special choice of filter parametrization.

%Note that while classical CNNs can be speed-up by leveraging the Fourier basis
%(instead of computing the convolutions in the spatial domain)
%\cite{mathieu_fast_2013}, it is only advantageous because of the $\bO(n \log(n))$
%complexity of the FFT, which has no equivalent on graphs (yet). 

\paragraph{Graph Fourier Transform.} We are interested in processing signals
defined on undirected and connected graphs $\G=(\V,\E,W)$, where $\V$ is a
finite set of $|\V|=n$ vertices, $\E$ is a set of edges and $W \in \R^{n \times n}$ is a
weighted adjacency matrix encoding the connection weight between two vertices. A
signal $x: \V \rightarrow \R$ defined on the nodes of the graph may be regarded
as a vector $x \in \R^n$ where $x_i$ is the value of $x$ at the $i^{th}$ node.
An essential operator in spectral graph analysis is the graph Laplacian
\cite{book:Chung97Spectral}, which combinatorial definition is $L = D - W \in
\R^{n \times n}$ where $D \in \R^{n \times n}$ is the diagonal degree matrix
with $D_{ii} = \sum_j W_{ij}$, and normalized definition is $L = I_n -
D^{-1/2} W D^{-1/2}$ where $I_n$ is the identity matrix. As $L$ is a real
symmetric positive semidefinite matrix, it has a complete set of orthonormal
eigenvectors $\{u_l\}_{l=0}^{n-1} \in \R^n$, known as the graph Fourier modes,
and their associated ordered real nonnegative eigenvalues
$\{\lambda_l\}_{l=0}^{n-1}$, identified as the frequencies of the graph. The
Laplacian is indeed diagonalized by the Fourier basis $U=[u_0, \ldots, u_{n-1}]
\in \R^{n \times n}$ such that $L = U \Lambda U^T$ where $\Lambda =
\diag([\lambda_0, \ldots, \lambda_{n-1}]) \in \R^{n \times n}$.
The graph Fourier transform of a signal $x \in \R^n$ is then defined as
$\hat{x} = U^T x \in \R^n$, and its inverse as $x = U \hat{x}$
\cite{art:ShumanNarangFrossardOrtegaVandergheynst13ReviewSPG}. As on Euclidean
spaces, that transform enables the formulation of fundamental
operations such as filtering.

\paragraph{Spectral filtering of graph signals.} As we cannot express a
meaningful translation operator in the vertex domain, the convolution operator
on graph $\ast_\G$ is defined in the Fourier domain such that $x \,\ast_\G\, y =
U((U^Tx) \odot (U^Ty))$, where $\odot$ is the element-wise Hadamard product. It
follows that a signal $x$ is filtered by $g_\theta$ as
\begin{equation}
	y = g_\theta(L) x = g_\theta(U \Lambda U^T) x = U g_\theta(\Lambda) U^T x.
\end{equation}
A non-parametric filter, i.e. a filter whose parameters
are all free, would be defined as
\begin{equation} \label{eq:filt_non-param}
	g_\theta(\Lambda) = \diag(\theta),
\end{equation}
where the parameter $\theta \in \R^n$ is a vector of Fourier coefficients.

\paragraph{Polynomial parametrization for localized filters.} There are however
two limitations with non-parametric filters: (i) they are not localized in
space and (ii) their learning complexity is in $\bO(n)$, the dimensionality of
the data. These issues can be overcome with the use of a polynomial filter
% a Laplacian-based polynomial spectral filter
\begin{equation} \label{eq:filt_poly}
	g_\theta(\Lambda) = \sum_{k=0}^{K-1} \theta_k \Lambda^k,
\end{equation}
where the parameter $\theta \in \R^K$ is a vector of polynomial coefficients.
The value at vertex $j$ of the filter $g_\theta$ centered at vertex $i$ is
given by $(g_\theta(L) \delta_i)_j = (g_\theta(L))_{i,j} = \sum_k \theta_k
(L^k)_{i,j}$, where the kernel is localized via a convolution with a Kronecker
delta function $\delta_i \in \R^n$. By \cite[Lemma
5.2]{art:HammondVandergheynstGribonval11GraphWav}, $d_\G(i,j) > K$ implies
$(L^K)_{i,j} = 0$, where $d_\G$ is the shortest path distance, i.e. the minimum
number of edges connecting two vertices on the graph.  Consequently, spectral
filters represented by $K^\text{th}$-order polynomials of the Laplacian are
exactly $K$-localized. Besides, their learning complexity is $\bO(K)$, the
support size of the filter, and thus the same complexity as classical CNNs.

% While there is many possible choice of recursive polynomials
\paragraph{Recursive formulation for fast filtering.} While we have shown how
to learn localized filters with $K$ parameters, the cost to filter a signal $x$
as $y = Ug_\theta(\Lambda)U^Tx$ is still high with $\bO(n^2)$ operations
because of the multiplication with the Fourier basis $U$. A solution to this
problem is to parametrize $g_\theta(L)$ as a polynomial function that can be
computed recursively from $L$, as $K$ multiplications by a sparse $L$ costs
$\bO(K|\E|) \ll \bO(n^2)$. One such polynomial, traditionally used in GSP to
approximate kernels (like wavelets), is the Chebyshev expansion
\cite{art:HammondVandergheynstGribonval11GraphWav}. Another option, the Lanczos
algorithm \cite{art:SusnjaraPerraudinKressnerVandergheynst15Lanczos}, which
constructs an orthonormal basis of the Krylov subspace $\mathcal{K}_K(L,x) =
\spn\{x,Lx,\ldots,L^{K-1}x\}$, seems attractive because of the coefficients'
independence. It is however more convoluted and thus left as a future work.

Recall that the Chebyshev polynomial $T_k(x)$ of order $k$ may be computed by
the stable recurrence relation $T_k(x) = 2x T_{k-1}(x) - T_{k-2}(x)$ with $T_0 =
1$ and $T_1 = x$. These polynomials form an orthogonal basis for $L^2([-1,1], dy
/ \sqrt{1-y^2})$, the Hilbert space of square integrable functions with respect
to the measure $dy/\sqrt{1-y^2}$. A filter can thus be parametrized as the
truncated expansion
\begin{equation} \label{eq:filt_cheby}
	g_\theta(\Lambda) = \sum_{k=0}^{K-1} \theta_k T_k(\tilde{\Lambda}),
\end{equation}
of order $K-1$,
where the parameter $\theta \in \R^K$ is a vector of Chebyshev coefficients and
$T_k(\tilde{\Lambda}) \in \R^{n \times n}$ is the Chebyshev polynomial of order
$k$ evaluated at $\tilde{\Lambda} = 2 \Lambda / \lambda_{max} - I_n$, a diagonal
matrix of scaled eigenvalues that lie in $[-1,1]$.
The filtering operation can then be written as $y = g_\theta(L) x
= \sum_{k=0}^{K-1} \theta_k T_k(\tilde{L}) x$, where $T_k(\tilde{L}) \in \R^{n
\times n}$ is the Chebyshev polynomial of order $k$ evaluated at the scaled
Laplacian $\tilde{L} = 2 L / \lambda_{max} - I_n$. %Note that the spectrum of the
%normalized Laplacian is bounded by $2$ \cite{book:Chung97Spectral}, such that
%the scaling can simply be $\tilde{L} = L - I_n$. 
Denoting $\bar{x}_k =
T_k(\tilde{L})x \in \R^n$, we can use the recurrence relation to compute
$\bar{x}_k = 2\tilde{L} \bar{x}_{k-1} - \bar{x}_{k-2}$ with $\bar{x}_0 = x$ and
$\bar{x}_1 = \tilde{L}x$. The entire filtering operation
	$y = g_\theta(L) x = [\bar{x}_0, \ldots, \bar{x}_{K-1}] \theta$
then costs $\bO(K|\E|)$ operations.

\paragraph{Learning filters.} The $j^\text{th}$ output feature map of the sample
$s$ is given by
\begin{equation} \label{eq:filterbank}
	y_{s,j} = \sum_{i=1}^{F_{in}} g_{\theta_{i,j}}(L) x_{s,i} \in \R^n,
\end{equation}
where the $x_{s,i}$ are the input feature maps and the $F_{in} \times F_{out}$
vectors of Chebyshev coefficients $\theta_{i,j} \in \R^K$ are the layer's
trainable parameters. When training multiple convolutional layers with the
backpropagation algorithm, one needs the two gradients
\begin{align}
	\frac{\partial E}{\partial \theta_{i,j}} =
	\sum_{s=1}^S [\bar{x}_{s,i,0}, \ldots, \bar{x}_{s,i,K-1}]^T
	\frac{\partial E}{\partial y_{s,j}}
	&& \text{and} &&
	\frac{\partial E}{\partial x_{s,i}} =
	\sum_{j=1}^{F_{out}} g_{\theta_{i,j}}(L)
	\frac{\partial E}{\partial y_{s,j}},
\end{align}
where $E$ is the loss energy over a mini-batch of $S$ samples. Each of the
above three computations boils down to $K$ sparse matrix-vector multiplications
and one dense matrix-vector multiplication for a cost of $\bO(K |\E| F_{in}
F_{out} S)$ operations. These can be efficiently computed on parallel
architectures by leveraging tensor operations. Eventually, $[\bar{x}_{s,i,0},
\ldots, \bar{x}_{s,i,K-1}]$ only needs to be computed once.

\subsection{Graph Coarsening} \label{sec:coarsening}

% Local stationarity property of data is extracted via localized convolutional
% kernels. We are now interested to extract the multi-scale hierarchical
% composition property of data. In standard CNNs, this is efficiently achieved
% via grid subsampling and pooling, which trades spatial resolution with
% feature resolution reducing the learning complexity without compromising the
% system performances. In contrast with regular domains, the subsampling
% operation on graphs or graph coarsening is not mathematically sound.

The pooling operation requires meaningful neighborhoods on graphs, where
similar vertices are clustered together. Doing this for multiple layers is
equivalent to a multi-scale clustering of the graph that preserves local
geometric structures. It is however known that graph clustering is
\mbox{NP-hard} \cite{art:BuiJonesGraphPartNPhard} and that approximations must
be used. While there exist many clustering techniques, e.g. the popular
spectral clustering \cite{art:VonLuxburg07Tutorial}, we are most interested in
multilevel clustering algorithms where each level produces a coarser graph
which corresponds to the data domain seen at a different resolution.  Moreover,
clustering techniques that reduce the size of the graph by a factor two at each
level offers a precise control on the coarsening and pooling size.  In this
work, we make use of the coarsening phase of the Graclus multilevel clustering
algorithm \cite{art:DhillonGuanKulis07Graclus}, which has been shown to be
extremely efficient at clustering a large variety of graphs. Algebraic
multigrid techniques on graphs \cite{art:RonSafroBrandt11MultigridGraph} and
the Kron reduction \cite{art:ShumanFarajiVandergheynst16PyramTrans} are two
methods worth exploring in future works.
%The later has the main advantage to preserve the ordering of the Laplacian
%spectrum, and may be able to commute with the graph convolution operator.
%\todo{If proven, this would be a very essential mathematical property of CNNs
%on graphs, exactly as downsampling for classical CNNs.}

Graclus \cite{art:DhillonGuanKulis07Graclus}, built on Metis
\cite{art:KarypisKumar98Metis}, uses a greedy algorithm to compute successive
coarser versions of a given graph and is able to minimize several popular
spectral clustering objectives, from which we chose the normalized cut
\cite{art:ShiMalik00NCut}. Graclus'
greedy rule consists, at each coarsening level, in picking an unmarked
vertex $i$ and matching it with one of its unmarked neighbors $j$ that
maximizes the local normalized cut $W_{ij} (1/d_i + 1/d_j)$. The two matched
vertices are then marked and the coarsened weights are set as the sum of their
weights. The matching is repeated until all nodes have been explored. This is
an very fast coarsening scheme which divides the number of nodes by
approximately two (there may exist a few singletons, non-matched nodes) from
one level to the next coarser level.

\subsection{Fast Pooling of Graph Signals}  \label{sec:pooling}

\begin{figure}[t]
\centering
\includegraphics[width=\textwidth]{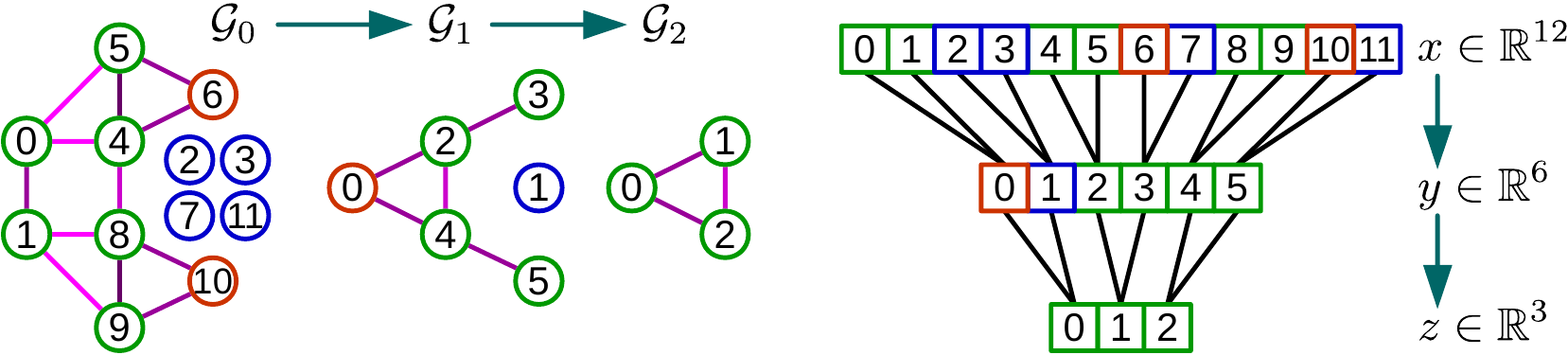}
\caption{\textbf{Example of Graph Coarsening and Pooling.} Let us carry out a max pooling of size 4 (or two poolings of size 2) on a signal $x \in
\R^8$ living on $\G_0$, the finest graph given as input. Note that it originally
possesses $n_0 = |\V_0| = 8$ vertices, arbitrarily ordered. For a pooling of
size 4, two coarsenings of size 2 are needed: let Graclus gives $\G_1$ of size
$n_1 = |\V_1| = 5$, then $\G_2$ of size $n_2 = |\V_2| = 3$, the coarsest graph.
Sizes are thus set to $n_2 = 3$, $n_1 = 6$, $n_0 = 12$ and fake nodes (in blue)
are added to $\V_1$ (1 node) and $\V_0$ (4 nodes) to pair with the singeltons
(in orange), such that each node has exactly two children.  Nodes in $\V_2$ are
then arbitrarily ordered and nodes in $\V_1$ and $\V_0$ are ordered
consequently.  At that point the arrangement of vertices in $\V_0$ permits a
regular 1D pooling on $x \in \R^{12}$ such that $z = [\max(x_0,x_1),
\max(x_4,x_5,x_6), \max(x_8,x_9,x_{10})] \in \R^3$, where the signal components
$x_2,x_3,x_7,x_{11}$ are set to a neutral value.}
\label{fig:pooling}
\end{figure}

Pooling operations are carried out many times and must be efficient. After
coarsening, the vertices of the input graph and its coarsened versions are not
arranged in any meaningful way. Hence, a direct application of the pooling
operation would need a table to store all matched vertices. That would result
in a memory inefficient, slow, and hardly parallelizable implementation. It is
however possible to arrange the vertices such that a graph
pooling operation becomes as efficient as a 1D pooling. We proceed in two
steps: (i) create a balanced binary tree and (ii) rearrange the vertices.
After coarsening, each node has either two children, if it was matched at the
finer level, or one, if it was not, i.e the node was a singleton. From the
coarsest to finest level, fake nodes, i.e.  disconnected nodes, are added to
pair with the singletons such that each node has two children. This structure
is a balanced binary tree: (i) regular nodes (and singletons) either have two
regular nodes (e.g. level 1 vertex 0 in \figref{pooling}) or (ii) one singleton
and a fake node as children (e.g. level 2 vertex 0), and
(iii) fake nodes always have two fake nodes as children (e.g. level 1 vertex 1).
Input signals are initialized with a neutral value at the
fake nodes, e.g. 0 when using a ReLU activation with max pooling. Because these
nodes are disconnected, filtering does not impact the initial neutral value.
While those fake nodes do artificially increase the dimensionality thus the
computational cost, we found that, in practice, the number of singletons left
by Graclus is quite low.
% < 10%? compared to the number of vertices.
Arbitrarily ordering the nodes at the coarsest level, then propagating this
ordering to the finest levels, i.e. node $k$ has nodes $2k$ and $2k+1$ as
children, produces a regular ordering in the finest level. Regular in the sense
that adjacent nodes are hierarchically merged at coarser levels. Pooling such a
rearranged graph signal is analog to pooling a regular 1D signal.
\figref{pooling} shows an example of the whole process.
This regular arrangement makes the operation very efficient and
satisfies parallel architectures such as GPUs as memory accesses are local,
i.e.  matched nodes do not have to be fetched.

\section{Related Works}

\subsection{Graph Signal Processing}

The emerging field of GSP aims at bridging the gap between signal processing
and spectral graph theory \cite{book:Chung97Spectral,
art:BelkinNiyogi05LaplaBeltrami, art:VonLuxburg07Tutorial}, a blend between
graph theory and harmonic analysis. A goal is to generalize fundamental
analysis operations for signals from regular grids to irregular structures
embodied by graphs. We refer the reader to
\cite{art:ShumanNarangFrossardOrtegaVandergheynst13ReviewSPG} for an
introduction of the field.  Standard operations on grids such as convolution,
translation, filtering, dilatation, modulation or downsampling do not extend
directly to graphs and thus require new mathematical definitions while keeping
the original intuitive concepts. In this context, the authors of
\cite{art:HammondVandergheynstGribonval11GraphWav, art:CoifmanLafon06DifMap,
pro:GavishNadlerCoifman10GraphHaar} revisited the construction of wavelet
operators on graphs and techniques to perform mutli-scale pyramid transforms on
graphs were proposed in \cite{art:ShumanFarajiVandergheynst16PyramTrans,
art:RamEladCohen11TreeWavelets}. The works of
\cite{pro:TsitsveroBarbarossa15Uncert, pro:PasdeloupAlamiGriponRabbat15Uncert,
art:PerraudinRicaudShumanVandergheynst16Uncert} redefined uncertainty
principles on graphs and showed that while intuitive concepts may be lost,
enhanced localization principles can be derived.

%In \cite{pro:HammondRaoaroorJacquesVandergheynst10LassoGraWav}, it was shown
%how to carry out lasso-based signal regularization on graphs, and studied the
%intertwined relationships between smoothness and sparsity on graphs. In
%\cite{pro:TremblayPuyGribonvalVandergheynst16CompSpecClus}, the authors
%investigated compressed sensing recovery conditions for graph spectral signal
%analysis.

\subsection{CNNs on Non-Euclidean Domains}

%Extending the success of CNNs to
%non-Euclidean domains is essential to boost the power analysis of data lying on
%complex networks like biological, social or telecommunication networks.

The Graph Neural Network framework \cite{scarselli_gnn_2009}, simplified in
\cite{li_ggsnn_2015}, was designed to embed each node in an Euclidean space
with a RNN and use those embeddings as features for classification or
regression of nodes or graphs. By setting their \textit{transition function}
$f$ as a simple diffusion instead of a neural net with a recursive relation,
their \textit{state} vector becomes $s = f(x) = Wx$.  Their point-wise
\textit{output function} $g_\theta$ can further be set as $\hat{x} =
g_\theta(s, x) = \theta (s - Dx) + x = \theta Lx + x$ instead of another neural
net. The Chebyshev polynomials of degree $K$ can then be obtained with a
$K$-layer GNN, to be followed by a non-linear layer and a graph pooling
operation. Our model can thus be interpreted as multiple layers of diffusions
and node-local operations.

The works of \cite{pro:GregorLeCun10LRF, pro:CoatesNg11LRF} introduced the
concept of constructing a local receptive field to reduce the number of learned
parameters. The idea is to group together features based upon a measure of
similarity such as to select a limited number of connections between two
successive layers. While this model reduces the number of parameters by
exploiting the locality assumption, it did not attempt to exploit any
stationarity property, i.e. no weight-sharing strategy.
%This approach is however not able to extract similar groups over the data
%domain, as it is not clear how to define convolutional filters on these
%groups.
The authors of \cite{art:BrunaZarembaSzlamLeCun13DLgraphs} used this idea for
their spatial formulation of graph CNNs. They use a weighted graph to define
the local neighborhood and compute a multiscale clustering of the graph for the
pooling operation. Inducing weight sharing in a spatial construction is however
challenging, as it requires to select and order the neighborhoods when a
problem-specific ordering (spatial, temporal, or otherwise) is missing.

A spatial generalization of CNNs to 3D-meshes, a class of smooth
low-dimensional non-Euclidean spaces, was proposed in \cite{masci2015geodesic}.
The authors used geodesic polar coordinates to define the convolution on
mesh patches, and formulated a deep learning architecture which allows
comparison across different manifolds. They obtained state-of-the-art results
for 3D shape recognition.

% For learning similarity graph
%In \cite{pro:ChenChengMallat14deepHaar} and \cite{pro:RustamovGuibas14deepHaar},
%the authors investigated the construction of Haar wavelet transforms on graphs
%using a deep hierarchical architecture. They applied the method to object
%recognition on sphere, and to sparse reconstruction of faces.
%\todo{Learn parameters of a wavelet kernel? Then it is a parametric method,
%they impose a shape to the filters.}
%\todo{Unknown graph geometry?}

The first spectral formulation of a graph CNN, proposed in
\cite{art:BrunaZarembaSzlamLeCun13DLgraphs}, defines a filter as
\begin{equation} \label{eq:filt_spline}
	g_\theta(\Lambda) = B \theta,
\end{equation}
where $B \in \R^{n \times K}$ is the cubic B-spline basis and the parameter
$\theta \in \R^K$ is a vector of control points.
They later proposed a strategy to learn the graph structure from the data and
applied the model to image recognition, text categorization and bioinformatics
\cite{art:HenaffBrunaLeCun15DLgraphs}.
This approach does however not scale up due to the necessary multiplications by
the graph Fourier basis $U$. Despite the cost of computing this matrix, which
requires an EVD on the graph Laplacian, the dominant cost is the need to
multiply the data by this matrix twice (forward and inverse Fourier transforms)
at a cost of $\bO(n^2)$ operations per forward and backward pass, a
computational bottleneck already identified by the authors. Besides, as they
rely on smoothness in the Fourier domain, via the spline parametrization, to
bring localization in the vertex domain, their model does not provide a precise
control over the local support of their kernels, which is essential to learn
localized filters. Our technique leverages on this work, and we showed how to
overcome these limitations and beyond.

\section{Numerical Experiments}

%All experiments were performed with TensorFlow, an open-source library for
%numerical computation using data flow graphs, especially suited for deep
%learning \cite{abadi_tensorflow_2016}. It features various backends, notably
%CUDA to compute on Nvidia GPUs.
%All computations are carried on an Nvidia Tesla K40c GPU.

In the sequel, we refer to the non-parametric and non-localized filters
\eqnref{filt_non-param} as \textit{Non-Param}, the filters \eqnref{filt_spline}
proposed in \cite{art:BrunaZarembaSzlamLeCun13DLgraphs} as \textit{Spline} and the proposed filters
\eqnref{filt_cheby} as \textit{Chebyshev}. We always use the Graclus coarsening
algorithm introduced in \secref{coarsening} rather than the simple
agglomerative method of \cite{art:BrunaZarembaSzlamLeCun13DLgraphs}. Our motivation is to compare the learned
filters, not the coarsening algorithms.

We use the following notation when describing network architectures: FC$k$
denotes a fully connected layer with $k$ hidden units, P$k$ denotes a (graph or
classical) pooling layer of size and stride $k$, GC$k$ and C$k$ denote a
(graph) convolutional layer with $k$ feature maps.
All FC$k$, C$k$ and GC$k$ layers are followed by a ReLU activation $\max(x,0)$.
The final layer is always a softmax regression and the loss energy $E$ is the
cross-entropy with an $\ell_2$ regularization on the weights of all FC$k$
layers. Mini-batches are of size $S = 100$.

\subsection{Revisiting Classical CNNs on MNIST} \label{sec:MNIST}
%\subsection{Comparison with classical CNNs on MNIST}
% Struct: motivation, dataset, model, results

\begin{table*}[t]
\centering
\begin{tabular}{llc}
\toprule
Model & Architecture & Accuracy \\
\midrule
%Linear SVM & 91.76  \\
%Softmax & 92.36  \\
Classical CNN & C32-P4-C64-P4-FC512 & 99.33  \\
Proposed graph CNN & GC32-P4-GC64-P4-FC512 & 99.14  \\
\bottomrule
\end{tabular}
\caption{Classification accuracies of the proposed graph CNN and a classical
CNN on MNIST.} 
\label{tab:mnist}
\end{table*}

To validate our model, we applied it to the Euclidean case on the benchmark
MNIST classification problem \cite{pro:LeCunBottouBengioHaffner98MNIST}, a
dataset of 70,000 digits represented on a 2D grid of size $28 \times 28$. For
our graph model, we construct an $8$-NN graph of the 2D grid which produces a
graph of $n = |\V| = 976$ nodes ($28^2 = 784$ pixels and 192 fake nodes as
explained in \secref{pooling}) and $|\E| = 3198$ edges. Following standard
practice, the weights of a $k$-NN similarity graph (between features) are
computed as
%$W_{ij} = \exp(-\|z_i - z_j\|_2^2 / \sigma^2)$
\begin{equation} \label{eq:knngraph}
	W_{ij} = \exp \left( -\frac{\|z_i - z_j\|_2^2}{\sigma^2} \right),
\end{equation}
where $z_i$ is the 2D coordinate of pixel $i$.
% and \sigma is set using the scheme proposed in

This is an important sanity check for our model, which must be able to extract
features on any graph, including the regular 2D grid. \tabref{mnist} shows the
ability of our model to achieve a performance very close to a classical CNN
with the same architecture. The gap in performance may be explained by the
isotropic nature of the spectral filters, i.e. the fact that edges in a general
graph do not possess an orientation (like up, down, right and left for pixels
on a 2D grid). Whether this is a limitation or an advantage depends on the
problem and should be verified, as for any invariance. Moreover, rotational
invariance has been sought: (i) many data augmentation schemes have used
rotated versions of images and (ii) models have been developed to learn this
invariance, like the Spatial Transformer Networks \cite{jaderberg2015spatial}.
Other explanations are the lack of experience on architecture design and the
need to investigate better suited optimization or initialization strategies.

The LeNet-5-like network architecture and the following hyper-parameters are
borrowed from the TensorFlow MNIST tutorial\footnote{
\url{https://www.tensorflow.org/versions/r0.8/tutorials/mnist/pros}}: dropout
probability of 0.5, regularization weight of $5\times10^{-4}$, initial learning
rate of 0.03, learning rate decay of 0.95, momentum of 0.9. Filters are of size
$5 \times 5$ and graph filters have the same support of $K = 25$. All models
were trained for 20 epochs.

\subsection{Text Categorization on 20NEWS}

\begin{table*}[t]
\begin{minipage}[b]{0.42\linewidth}
	\centering
	\begin{tabular}[b]{lc}
		\toprule
		Model & Accuracy \\
		\midrule
		Linear SVM & 65.90 \\
		Multinomial Naive Bayes & 68.51 \\
		%Multinomial logistic regression (softmax) & 66.28 \\
		Softmax & 66.28 \\
		\addlinespace
		FC2500 & 64.64 \\
		FC2500-FC500 & 65.76 \\
		\addlinespace
		GC32 & 68.26 \\
		%GC32-FC500 & \todo{?} \\
		\bottomrule
	\end{tabular}
	\caption{Accuracies of the proposed graph CNN and other
	methods on 20NEWS.} 
	\label{tab:20news}
\end{minipage}
\hfill
\begin{minipage}[b]{0.55\linewidth}
	\centering
	\includegraphics[width=\textwidth]{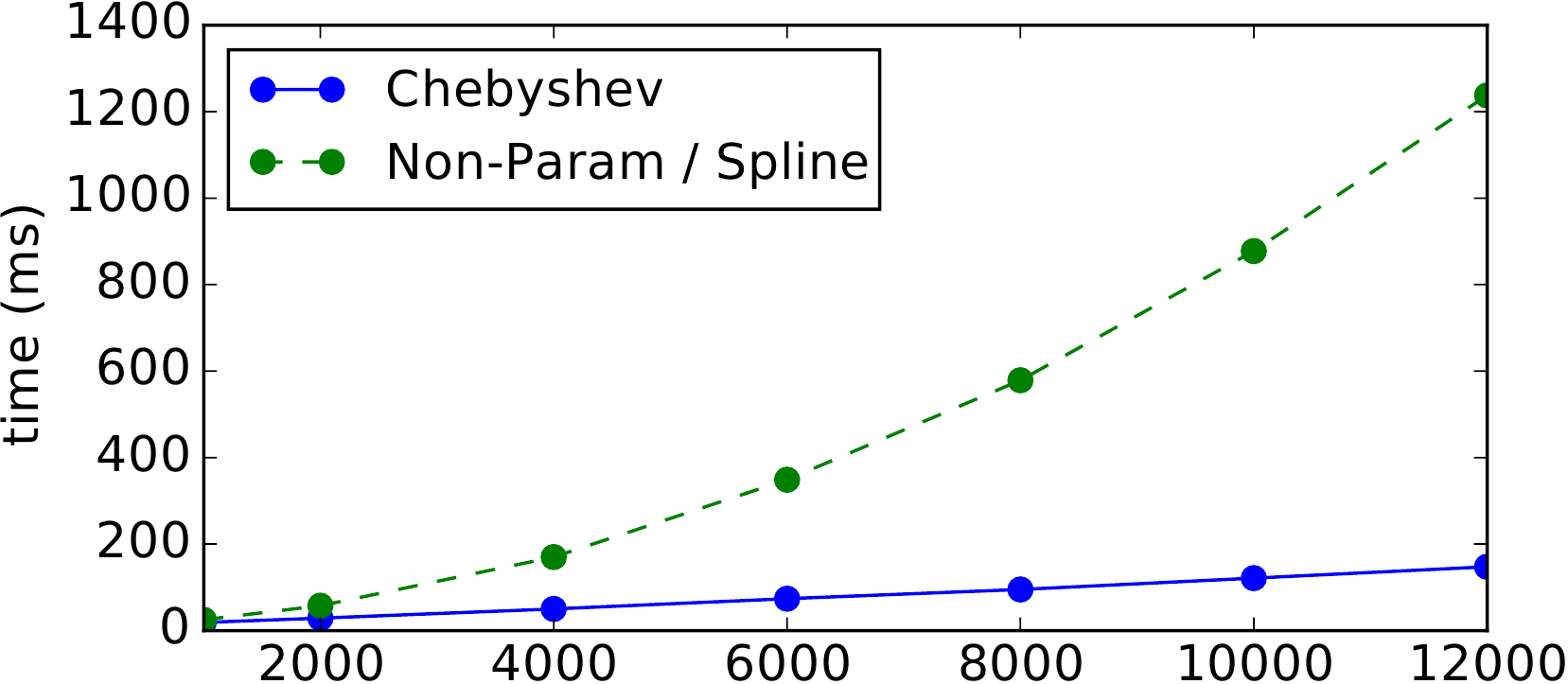}
	\captionof{figure}{Time to process a mini-batch of $S=100$ 20NEWS documents
	w.r.t. the number of words $n$.}
	\label{fig:runtime}
\end{minipage}
\end{table*}

To demonstrate the versatility of our model to work with graphs generated from
unstructured data, we applied our technique to the text categorization problem
on the 20NEWS dataset which consists of 18,846 (11,314 for training and 7,532
for testing) text documents associated with 20 classes
\cite{art:Joachims9620NEWS}. We extracted the 10,000 most common words from the
93,953 unique words in this corpus. Each document $x$ is represented using the
bag-of-words model, normalized across words. To test our model, we constructed
a 16-NN graph with \eqnref{knngraph} where $z_i$ is the word2vec embedding
\cite{pro:MikolovChenCorradoDean13word2vec} of word $i$, which produced a graph
of $n = |\V| = 10,000$ nodes and $|\E| = 132,834$ edges. All models were
trained for 20 epochs by the Adam optimizer \cite{art:KingmaBa14AdamOpt} with
an initial learning rate of 0.001. The architecture is GC32 with support $K =
5$. \tabref{20news} shows decent performances: while the proposed model does
not outperform the multinomial naive Bayes classifier on this small dataset, it
does defeat fully connected networks, which require much more parameters.

%todo hyper-parameters

\subsection{Comparison between Spectral Filters and Computational Efficiency}

\begin{table*}[t!] \centering
\begin{tabular}{llccc} \toprule
& & \multicolumn{3}{c}{Accuracy} \\
\cmidrule{3-5}
Dataset & Architecture & Non-Param \eqnref{filt_non-param} &
Spline \eqnref{filt_spline} \cite{art:BrunaZarembaSzlamLeCun13DLgraphs} &
Chebyshev \eqnref{filt_cheby} \\
\midrule
MNIST & GC10 & 95.75 & 97.26 & {97.48} \\
MNIST & GC32-P4-GC64-P4-FC512 & 96.28 & 97.15 & {99.14} \\
%20NEWS & GC32 & \todo{?} & \todo{?} & \todo{?} \\
\bottomrule \end{tabular}
\caption{Classification accuracies for different types of spectral filters
($K=25$).} % \todo{on MNIST and $K=5$ on 20NEWS.}
\label{tab:filters}
\end{table*}

\begin{table*}[t]
\centering
%\vspace{-20pt}
\begin{tabular}{llccc} \toprule
& & \multicolumn{2}{c}{Time (ms)} & \\
Model & Architecture & CPU & GPU & Speedup \\ % {\small(Tesla K40c)} 
\midrule
Classical CNN & C32-P4-C64-P4-FC512 & 210 & 31 & 6.77x \\
Proposed graph CNN & GC32-P4-GC64-P4-FC512 & 1600 & 200 & 8.00x \\
\bottomrule \end{tabular}
\caption{Time to process a mini-batch of $S=100$ MNIST images.} 
\label{tab:speedup}
\end{table*}

\begin{figure}[t]
\centering
\includegraphics[width=\textwidth]{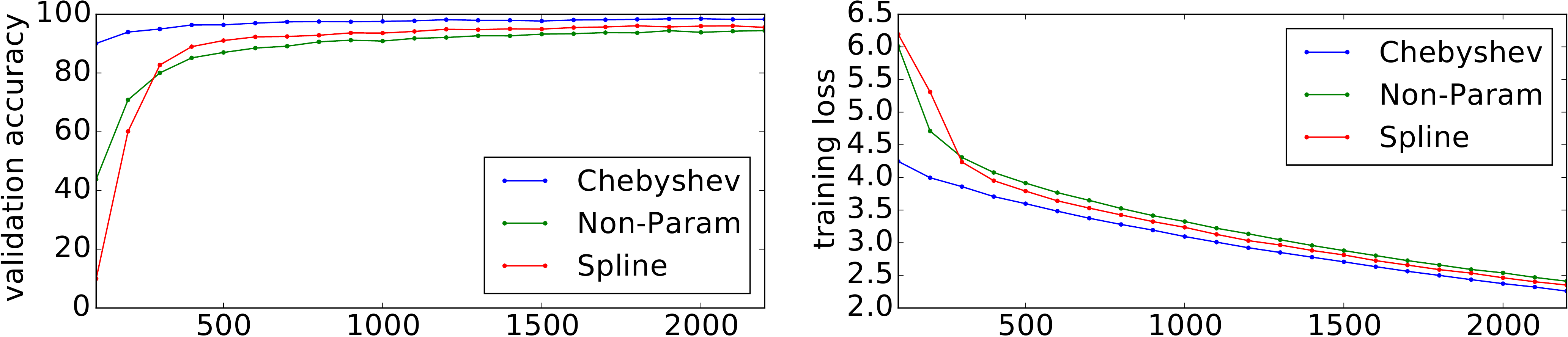}
\caption{Plots of validation accuracy and training loss for the first 2000
iterations on MNIST.}
\label{fig:convergence}
\end{figure}

\tabref{filters} reports that the proposed parametrization \eqnref{filt_cheby}
outperforms \eqnref{filt_spline} from \cite{art:BrunaZarembaSzlamLeCun13DLgraphs} as well as non-parametric
filters \eqnref{filt_non-param} which are not localized and require $\bO(n)$
parameters. Moreover, \figref{convergence} gives a sense of how the validation
accuracy and the loss $E$ converges w.r.t. the filter definitions.

\figref{runtime} validates the low computational complexity of our model which
scales as $\bO(n)$ while \cite{art:BrunaZarembaSzlamLeCun13DLgraphs} scales as $\bO(n^2)$. The measured
runtime is the total training time divided by the number of gradient steps.
\tabref{speedup} shows a similar speedup as classical CNNs when moving to GPUs.
This exemplifies the parallelization opportunity offered by our model, who
relies solely on matrix multiplications. Those are efficiently implemented by
cuBLAS, the linear algebra routines provided by NVIDIA.

\subsection{Influence of Graph Quality} \label{sec:graph_quality}

\begin{table*}[t]
\centering
\begin{tabular}{lcc} \toprule
Architecture & 8-NN on 2D Euclidean grid & random \\
\midrule
GC32 & 97.40 & 96.88 \\
GC32-P4-GC64-P4-FC512 & 99.14 & 95.39 \\
\bottomrule \end{tabular}
\caption{Classification accuracies with different graph constructions on MNIST.} 
\label{tab:mnist_quality}
\end{table*}

\begin{table*}[t] \centering
\begin{tabular}{ccccc} \toprule
& \multicolumn{2}{c}{word2vec} & & \\
bag-of-words & pre-learned & learned & approximate & random \\
\midrule
67.50 & 66.98 & 68.26 & 67.86 & 67.75 \\
\bottomrule \end{tabular}
\caption{Classification accuracies of GC32 with different graph constructions on 20NEWS.} 
\label{tab4b}
\label{tab:20news_quality}
\end{table*}

% While we know that these data properties are true for low-dimensional
% Euclidean data like audios, images and videos, we also show experimentally
% that they are also satisfied for text documents as long as the graph is
% properly constructed.

% More importantly, these results verify the validity of the statistical
% assumptions made on the data, that are locality and stationarity, and which
% are at the core of the design of any CNN technique. 

For any graph CNN to be successful, the statistical assumptions of locality,
stationarity, and compositionality regarding the data must be fulfilled on the
graph where the data resides. Therefore, the learned filters' quality and thus
the classification performance critically depends on the quality of the graph.
For data lying on Euclidean space, experiments in \secref{MNIST} show that a
simple $k$-NN graph of the grid is good enough to recover almost exactly the
performance of standard CNNs. We also noticed that the value of $k$ does not
have a strong influence on the results. We can witness the importance of a
graph satisfying the data assumptions by comparing its performance with a
random graph. \tabref{mnist_quality} reports a large drop of accuracy when
using a random graph, that is when the data structure is lost and the
convolutional layers are not useful anymore to extract meaningful features.

While images can be structured by a grid graph, a feature graph has to be built
for text documents represented as bag-of-words. We investigate here three ways
to represent a word $z$: the simplest option is to represent each word as its
corresponding column in the bag-of-words matrix while, another approach is to
learn an embedding for each word with word2vec
\cite{pro:MikolovChenCorradoDean13word2vec} or to use the pre-learned
embeddings provided by the authors. For larger datasets, an approximate nearest
neighbors algorithm may be required, which is the reason we tried LSHForest
\cite{pro:BawaCondieGanesan05LSHForest} on the learned word2vec embeddings.
\tabref{20news_quality} reports classification results which highlight the
importance of a well constructed graph.

%the graph built on the learned word2vec embedding is the
%best at capturing the local and stationarity properties of text documents. It is
%worth noticing that the approximate $k$-NN graph constructed from it is almost
%as bad as a random graph, meaning that it may be a better strategy to learn a
%good low-dimensional embedding first and then construct an exact $k$-NN graph
%from this embedding.

\section{Conclusion and Future Work}

In this paper, we have introduced the mathematical and computational
foundations of an efficient generalization of CNNs to graphs using tools from
GSP. Experiments have shown the ability of the model to extract local and
stationary features through graph convolutional layers. Compared with the first
work on spectral graph CNNs introduced in \cite{art:BrunaZarembaSzlamLeCun13DLgraphs}, our model provides a
strict control over the local support of filters, is computationally more
efficient by avoiding an explicit use of the Graph Fourier basis, and
experimentally shows a better test accuracy. Besides, we addressed the three
concerns raised by \cite{art:HenaffBrunaLeCun15DLgraphs}: (i) we introduced a
model whose computational complexity is linear with the dimensionality of the
data, (ii) we confirmed that the quality of the input graph is of paramount
importance, (iii) we showed that the statistical assumptions of local
stationarity and compositionality made by the model are verified for text
documents as long as the graph is well constructed.

Future works will investigate two directions. On one hand, we will enhance the
proposed framework with newly developed tools in GSP. On the other hand, we
will explore applications of this generic model to important fields where the
data naturally lies on graphs, which may then incorporate external information
about the structure of the data rather than artificially created graphs which
quality may vary as seen in the experiments. Another natural and future
approach, pioneered in \cite{art:HenaffBrunaLeCun15DLgraphs}, would be to
alternate the learning of the CNN parameters and the graph.

\clearpage
\bibliographystyle{plain}
%{\setstretch{0} \small
%\bibliography{bib_nips16}
%%\bibliography{bib_nips16_short}
%}

{
	\setlength{\parskip}{2.1pt}
	\setstretch{0}
	\small
	\bibliography{refs}

\begin{thebibliography}{10}

\bibitem{abadi_tensorflow_2016}
Martín Abadi~et al.
\newblock {{TensorFlow}}: {{Large-Scale Machine Learning}} on {{Heterogeneous
  Distributed Systems}}.
\newblock 2016.

\bibitem{pro:BawaCondieGanesan05LSHForest}
M.~Bawa, T.~Condie, and P.~Ganesan.
\newblock {LSH Forest: Self-Tuning Indexes for Similarity Search}.
\newblock In {\em International Conference on World Wide Web}, pages 651--660,
  2005.

\bibitem{art:BelkinNiyogi05LaplaBeltrami}
M.~Belkin and P.~Niyogi.
\newblock {Towards a Theoretical Foundation for Laplacian-based Manifold
  Methods}.
\newblock {\em Journal of Computer and System Sciences}, 74(8):1289--1308,
  2008.

\bibitem{art:BrunaZarembaSzlamLeCun13DLgraphs}
J.~Bruna, W.~Zaremba, A.~Szlam, and Y.~LeCun.
\newblock {Spectral Networks and Deep Locally Connected Networks on Graphs}.
\newblock {\em arXiv:1312.6203}, 2013.

\bibitem{art:BuiJonesGraphPartNPhard}
T.N. Bui and C.~Jones.
\newblock {Finding Good Approximate Vertex and Edge Partitions is NP-hard}.
\newblock {\em Information Processing Letters}, 42(3):153--159, 1992.

\bibitem{book:Chung97Spectral}
F.~R.~K. Chung.
\newblock {\em {Spectral Graph Theory}}, volume~92.
\newblock American Mathematical Society, 1997.

\bibitem{pro:CoatesNg11LRF}
A.~Coates and A.Y. Ng.
\newblock {Selecting Receptive Fields in Deep Networks}.
\newblock In {\em Neural Information Processing Systems (NIPS)}, pages
  2528--2536, 2011.

\bibitem{art:CoifmanLafon06DifMap}
R.R. Coifman and S.~Lafon.
\newblock {Diffusion Maps}.
\newblock {\em Applied and Computational Harmonic Analysis}, 21(1):5--30, 2006.

\bibitem{art:DhillonGuanKulis07Graclus}
I.~Dhillon, Y.~Guan, and B.~Kulis.
\newblock {Weighted Graph Cuts Without Eigenvectors: A Multilevel Approach}.
\newblock {\em IEEE Transactions on Pattern Analysis and Machine Intelligence
  (PAMI)}, 29(11):1944--1957, 2007.

\bibitem{pro:GavishNadlerCoifman10GraphHaar}
M.~Gavish, B.~Nadler, and R.~Coifman.
\newblock {Multiscale Wavelets on Trees, Graphs and High Dimensional Data:
  Theory and Applications to Semi Supervised Learning}.
\newblock In {\em International Conference on Machine Learning (ICML)}, pages
  367--374, 2010.

\bibitem{pro:GregorLeCun10LRF}
K.~Gregor and Y.~LeCun.
\newblock {Emergence of Complex-like Cells in a Temporal Product Network with
  Local Receptive Fields}.
\newblock In {\em arXiv:1006.0448}, 2010.

\bibitem{art:HammondVandergheynstGribonval11GraphWav}
D.~Hammond, P.~Vandergheynst, and R.~Gribonval.
\newblock {Wavelets on Graphs via Spectral Graph Theory}.
\newblock {\em Applied and Computational Harmonic Analysis}, 30(2):129--150,
  2011.

\bibitem{art:HenaffBrunaLeCun15DLgraphs}
M.~Henaff, J.~Bruna, and Y.~LeCun.
\newblock {Deep Convolutional Networks on Graph-Structured Data}.
\newblock {\em arXiv:1506.05163}, 2015.

\bibitem{jaderberg2015spatial}
Max Jaderberg, Karen Simonyan, Andrew Zisserman, et~al.
\newblock Spatial transformer networks.
\newblock In {\em Advances in Neural Information Processing Systems}, pages
  2017--2025, 2015.

\bibitem{art:Joachims9620NEWS}
T.~Joachims.
\newblock {A Probabilistic Analysis of the Rocchio Algorithm with TFIDF for
  Text Categorization}.
\newblock {\em Carnegie Mellon University, Computer Science Technical Report},
  CMU-CS-96-118, 1996.

\bibitem{art:KarypisKumar98Metis}
G.~Karypis and V.~Kumar.
\newblock {A Fast and High Quality Multilevel Scheme for Partitioning Irregular
  Graphs}.
\newblock {\em SIAM Journal on Scientific Computing (SISC)}, 20(1):359--392,
  1998.

\bibitem{art:KingmaBa14AdamOpt}
D.~Kingma and J.~Ba.
\newblock {Adam: A Method for Stochastic Optimization}.
\newblock {\em arXiv:1412.6980}, 2014.

\bibitem{art:LeCunBengioHinton15DL}
Y.~LeCun, Y.~Bengio, and G.~Hinton.
\newblock {Deep Learning}.
\newblock {\em Nature}, 521(7553):436--444, 2015.

\bibitem{pro:LeCunBottouBengioHaffner98MNIST}
Y.~LeCun, L.~Bottou, Y.~Bengio, and P.~Haffner.
\newblock {Gradient-Based Learning Applied to Document Recognition}.
\newblock In {\em Proceedings of the IEEE, 86(11)}, pages 2278--2324, 1998.

\bibitem{li_ggsnn_2015}
Yujia Li, Daniel Tarlow, Marc Brockschmidt, and Richard Zemel.
\newblock Gated {{Graph Sequence Neural Networks}}.

\bibitem{art:VonLuxburg07Tutorial}
U.~Von Luxburg.
\newblock {A Tutorial on Spectral Clustering}.
\newblock {\em Statistics and Computing}, 17(4):395--416, 2007.

\bibitem{book:Mallat99wavelets}
S.~Mallat.
\newblock {\em {A Wavelet Tour of Signal Processing}}.
\newblock Academic press, 1999.

\bibitem{masci2015geodesic}
Jonathan Masci, Davide Boscaini, Michael Bronstein, and Pierre Vandergheynst.
\newblock Geodesic convolutional neural networks on riemannian manifolds.
\newblock In {\em Proceedings of the IEEE International Conference on Computer
  Vision Workshops}, pages 37--45, 2015.

\bibitem{pro:MikolovChenCorradoDean13word2vec}
T.~Mikolov, K.~Chen, G.~Corrado, and J.~Dean.
\newblock {Estimation of Word Representations in Vector Space}.
\newblock In {\em International Conference on Learning Representations}, 2013.

\bibitem{pro:PasdeloupAlamiGriponRabbat15Uncert}
B.~Pasdeloup, R.~Alami, V.~Gripon, and M.~Rabbat.
\newblock {Toward an Uncertainty Principle for Weighted Graphs}.
\newblock In {\em Signal Processing Conference (EUSIPCO)}, pages 1496--1500,
  2015.

\bibitem{art:PerraudinRicaudShumanVandergheynst16Uncert}
N.~Perraudin, B.~Ricaud, D.~Shuman, and P.~Vandergheynst.
\newblock {Global and Local Uncertainty Principles for Signals on Graphs}.
\newblock {\em arXiv:1603.03030}, 2016.

\bibitem{art:RamEladCohen11TreeWavelets}
I.~Ram, M.~Elad, and I.~Cohen.
\newblock {Generalized Tree-based Wavelet Transform}.
\newblock {\em IEEE Transactions on Signal Processing,}, 59(9):4199--4209,
  2011.

\bibitem{art:RonSafroBrandt11MultigridGraph}
D.~Ron, I.~Safro, and A.~Brandt.
\newblock {Relaxation-based Coarsening and Multiscale Graph Organization}.
\newblock {\em SIAM Iournal on Multiscale Modeling and Simulation}, 9:407--423,
  2011.

\bibitem{scarselli_gnn_2009}
F.~Scarselli, M.~Gori, A.~C. Tsoi, M.~Hagenbuchner, and G.~Monfardini.
\newblock The {{Graph Neural Network Model}}.
\newblock 20(1):61--80.

\bibitem{art:ShiMalik00NCut}
J.~Shi and J.~Malik.
\newblock {Normalized Cuts and Image Segmentation}.
\newblock {\em IEEE Transactions on Pattern Analysis and Machine Intelligence
  (PAMI)}, 22(8):888--905, 2000.

\bibitem{art:ShumanNarangFrossardOrtegaVandergheynst13ReviewSPG}
D.~Shuman, S.~Narang, P.~Frossard, A.~Ortega, and P.~Vandergheynst.
\newblock {The Emerging Field of Signal Processing on Graphs: Extending
  High-Dimensional Data Analysis to Networks and other Irregular Domains}.
\newblock {\em IEEE Signal Processing Magazine}, 30(3):83--98, 2013.

\bibitem{art:ShumanFarajiVandergheynst16PyramTrans}
D.I. Shuman, M.J. Faraji, and P.~Vandergheynst.
\newblock {A Multiscale Pyramid Transform for Graph Signals}.
\newblock {\em IEEE Transactions on Signal Processing}, 64(8):2119--2134, 2016.

\bibitem{art:SusnjaraPerraudinKressnerVandergheynst15Lanczos}
A.~Susnjara, N.~Perraudin, D.~Kressner, and P.~Vandergheynst.
\newblock {Accelerated Filtering on Graphs using Lanczos Method}.
\newblock {\em preprint arXiv:1509.04537}, 2015.

\bibitem{pro:TsitsveroBarbarossa15Uncert}
M.~Tsitsvero and S.~Barbarossa.
\newblock {On the Degrees of Freedom of Signals on Graphs}.
\newblock In {\em Signal Processing Conference (EUSIPCO)}, pages 1506--1510,
  2015.

\end{thebibliography}
}

\end{document}